# A novel approach of solving the CNF-SAT problem

**Abstract** – In this paper, we discussed CNF-SAT problem (NP-Complete problem) and analysis two solutions that can solve the problem, the PL-Resolution algorithm and the WalkSAT algorithm. PL-Resolution is a sound and complete algorithm that can be used to determine satisfiability and unsatisfiability with certainty. WalkSAT can determine satisfiability if it finds a model, but it cannot guarantee to find a model even there exists one. However, WalkSAT is much faster than PL-Resolution, which makes WalkSAT more practical; and we have analysis the performance between these two algorithms, and the performance of WalkSAT is acceptable if the problem is not so hard.

## 1. Introduction

The CNF Satisfiability Problem (CNF-SAT) is a version of the Satisfiability (SAT) Problem, where the Boolean formula is represented in the Conjunctive Normal Form (CNF).

### 1.1 Boolean Satisfiability

Boolean Satisfiability (SAT), in the Computer Science area, is a problem of determining whether there exists a Boolean value assignment, which satisfies the given formula. In other words, given a Boolean expression written using only AND ($\wedge$), OR ($\vee$), NOT ($\neg$), variables, and parentheses, is there some assignment of TRUE and FALSE values to the variables that make the entire expression evaluate to TRUE. On the other hand, the formula is unsatisfiable if no such assignment exists, which indicate that the entire expression is evaluated to FALSE for all possible variables assignments. For example, the Boolean expression $A \wedge B$, setting the both variables A and B to be TRUE to make the whole sentence TRUE, i.e., the expression $A \wedge B$ is satisfiable. However, there exists unsatisfiable expression, such as $A \wedge \neg A$, no matter variable A is TRUE or FALSE, the expression $A \wedge \neg A$ is always FALSE. The SAT problem can be applied in many fields of Computer Science and Engineering, including underlying model for a significant and increasing number of applications in Electronics Design Automation (EDA). Furthermore, a large number of problems that occur in knowledge base, learning, planning and other areas of areas of artificial intelligence (AI) are essentially the problems of SAT.

### 1.2 Conjunctive Normal Form:

Conjunctive Normal Form (CNF) is a formula, which is conjunction of clauses, where clause is a disjunction of literals; any formula satisfies the above conditions, we say the formula is in CNF; and it is a useful representation of Boolean expression, and useful in automated theorem proving. For instance, the formula $A \wedge (B \vee C)$ is in CNF. Furthermore, every propositional formula can be converted into an equivalent formula that is in CNF by using rules of logical equivalences. In addition, there are other forms such as Disjunctive Normal Form (DNF) and Horn Form. However, we will mainly focus on CNF here.

## 2. CNF-SAT Problem

### 2.1 Practical problem:

The CNF-SAT problem can be applied to solve many practical problems in the real world. Consider a real life problem, suppose you have a wedding to plan, and want to arrange the wedding seating for a certain number of guests in a hall. The hall has a certain number tables for seating. Some pairs of guests are couples or close Friends, and want to sit together at the same table. Some other pairs of guests are Enemies, and must be separated into different tables. The rest pairs are Indifferent with each other, and do not mind sitting together or not. However, each pair of guests can only have one relationship, Friends, Enemies or Indifferent. Then the problem is to find a seating arrangement that satisfies all the constraints.

### 2.2 Convert to CNF-SAT problem

This problem can be translated into symbolic logic, where each variable is TRUE or FALSE. Consider the example, we can first construct constraint a sentence which maybe close to our language and then convert to CNF. Then we can get the following constraint:

(a) Each guest i should be seated at least one table.
(b) Each guest i should be seated at most one table.
(c) Any two guests i and j who are Friends should be seated at the same table.
(d) Any two guests i and j who are Enemies should be seated at different tables.

Then we convert the above sentences into the formula in CNF. Suppose there are M guests in total, and there are N tables in the hall. The number of pairs of Friends is F, and the number of pairs of Enemies is E; Let X(i,n) denotes that guest i sit at table n, and we can get the corresponding formulas, and guest i and f are friends, i and e are enemies:

(a) $X(i, 1) \vee X(i, 2) \vee \ldots X(i, n) \vee \ldots X(i, N)$

$(1 \leq i \leq M, 1 \leq n \leq N)$

(b) $\neg X(i, k) \vee \neg X(i, n)$

$(1 \leq i \leq M, 1 \leq k \neq n \leq N)$

(c) $\{\neg X(i, n) \vee X(f, n)\} \wedge \{\neg X(f, n) \vee X(i, n)\}$

$(1 \leq i, f \leq M, 1 \leq n \leq N)$

(d) $\neg X(i, n) \vee \neg X(j, n)$

$(1 \leq i, j \leq M, 1 \leq n \leq N)$

Then the whole sentence which contain all the constraint should be the conjunction of all the above clauses:

$\{X(i, 1) \vee X(i, 2) \vee \ldots X(i, n) \vee \ldots X(i, N)\} \wedge \{\neg X(i, k) \vee \neg X(i, n)\} \wedge \{\neg X(i, n) \vee X(f, n)\} \wedge \{\neg X(f, n) \vee X(i, n)\} \wedge \{\neg X(i, n) \vee \neg X(j, n)\}$

$(1 \leq i \leq M, 1 \leq k \neq n \leq N)$

If we can find a Boolean value assignment for each term X(i, n) in the above sentence and make the sentence evaluate TRUE, then we solve our wedding seating arrangement problem.

**2.3 NP-Complete problem:**

CNF-SAT is a kind of NP-Complete (NPC) problem, which means we cannot find optimal solutions in polynomial time. NPC is a class of decision problem, i.e., if we already have a solution on hand, then we can check whether the solution is correct in polynomial time, just by substituting all the variables with the Boolean assignment and check whether the sentence is TRUE; however, finding such a Boolean assignment is in exponential time. Consider just using brute force to solve the above wedding seating arrangement problem, we try assigning all the possible values for the sentence, then the space would be $O(2^{MN})$, which grows exponentially with the increase of M and N. However, if we are using AI technology to find the solution more intelligently, we can find the solution more quickly but the solution is suboptimal, which we will discuss in detail later.

**3. Solutions and Analysis**

There already exists many SAT solvers such as Chaff, HyperSAT, Spear, The MiniSAT Solver, etc. We will discuss two solutions for CNF-SAT problem, PL-Resolution Algorithm and WalkSAT Algorithm.

**3.1 PL-Resolution Algorithm:**

The PL-Resolution Algorithm is a sound and complete solution that guarantees to determine whether there exists a solution or not. The main idea of this algorithm is achieve new clauses by resolving each pair of clauses in the Knowledge Base (KB), For example, clause $A \vee B$ and clause $\neg A \vee C$ can be resolved to a new clause $B \vee C$. Then, add the new clauses into KB. Only two results would occur in the end: (1) the resolve step gives us an empty clause, indicating there is a contradiction in KB, which means KB is unsatisfiable. For instance, clause A and ¬A would be resolved to empty clause, besides, we know $A \wedge \neg A$ is always FALSE intuitively; (2) otherwise the new clauses we gain are already in the KB, which mean there is no contradiction, and KB is satisfiable. The pseudo code of this algorithm [1] can be written as follow:

**function** PL-RESOLUTION(*KB*) **returns** *true* or *false*
   **inputs**: *KB*, the knowledge base, a sentence in propositional logic
   *clauses* <– the set of clauses in the CNF representation of *KB*
   *new* <– { }
   **loop do**
      for each pair of clauses *Ci*, *Cj* in clauses do
         *resolvents* <– PL-RESOLVE(*Ci*, *Cj*)
         if *resolvents* contains the empty clause **then return** *false*
         *new* <– *new* ∪ *resolvents*
      **if** *new* ∈ *clauses* then **return** *true*
      *clauses* <– *clauses* ∪ *new*

**3.2 WalkSAT Algorithm:**

Unlike PL-Resolution, the WalkSAT Algorithm can determine satisfiability (if it finds a model), but it cannot absolutely determine unsatisfiability. WalkSAT is one of the simplest and most effective algorithms. At first, WalkSAT randomly generate a model and assign all the variables with that model. If the sentence is unsatisfiable under this model, it will modify the model. It chooses a unsatisfied clause randomly and selects a variable in taht clause to flip in every iteration. There are two way of deciding which variable should be flipped: (1) "min-conflicts" step, which minimize the number of unsatisfied clauses in the new state; (2) "random walk" step, which just picks the variable randomly. The pseudo code of this algorithm[1] can be written as follow:

**function** WALKSAT(*clauses*, *p*, *max_flips*) **return** a satisfying *model* or *failure*
   **inputs**: *clauses*, a set of clauses in propositional logic
        *p*, the probability of choosing to do a "random walk" move, typically around 0.5

    *max_flips*, number of flips allowed before giving up
    *model* <− a random assignment of *true/false* to the symbols in *clauses*
     **for** *i* = 1 **to** *max_flips* **do**
      **if** *model* satisfies *clauses* then **return** *model*
      *clause* <− a randomly selected clause from *clauses* that is *false* in *model*
       **with probability** *p* flip the value in *model* of a randomly selected symbol from *clause*
       **else** flip whichever symbol in *clause* maximizes the number of satisfied clauses
     **return** *failure*

When WalkSAT returns a model, the input sentence is indeed satisfiable, and we also find the solution which is its return model, but when it returns failure, there are two possible causes: either the sentence is unsatisfiable or the number of iteration reaches the *max_flips*. The mechanism of this algorithm cannot guarantee to decide whether the input sentence is satisfiable or not. However, this algorithm is much faster compared to PL-Resolution and can gain good performance if we set *p* and *max_flips* properly.

## 4. Experiment

We will compare these two algorithms in terms of performance and use the example of seating arrangement problem that we mentioned before. In order to simulate this problem, we first generate randomly relationship for each pairs of guests, setting any two guests are Friends with probability *f*, or are Enemies with probability *e*. And any two guests are Indifferent to each other with probability *1-f-e*. Besides, we also set the number of guests to be *M* and the number of tables to be *N*. Then we convert our generated instance of wedding seating arrangement into a CNF sentence.

The difficulty of wedding seat arrangement problem mostly results from dealing with the pairs of Enemies among guests. We use both algorithms to produce a plot of *P*, which is the possibility of the instance is satisfiability, as a function of the probability *e* with which any two guests are Enemies. Suppose we have a small wedding to plan, and set *M*=16 and *N*=2. In order to eliminate the influence of Friends relationship, we set *f*=0. Generate a set of 100 random sentences for each setting of *e*, which increases from 2% to 20% at an interval of 2%, and use both algorithms to determine whether they are satisfiable. For WalkSAT, we set *p*=50% and *max_flips*=100. We plot the results of P versus *e* for both algorithms on the same graph as follow:

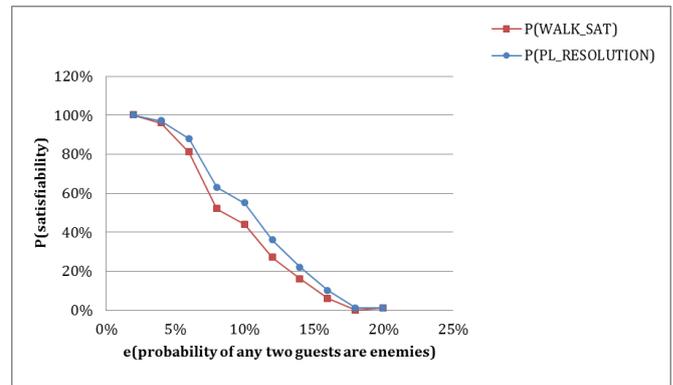

**Figure 1.** P versus e for WalkSAT and PL-Resolution.

According to Figure, we can see that the *P* of PL-Resolution is always higher than or equal to that of WalkSAT, which support the different ideal of PL-Resolution and WalkSAT. Though the performance of PL-Resolution is always better than WalkSAT, the runtime of PL-Resolution is growing exponentially. At the beginning, the WalkSAT performance as well as PL-Resolution did as we expected, because the problem is easy at first. As the *e* increasing, the problem become much more hard because the condition is more restrict. When *e* reach 20%, both algorithm can hardly find solution.

## 5. Conclusions

In this paper, we have discussed the CNF-SAT problem, which is a NP-Complete problem and analysis two solutions that can solve the problem, the PL-Resolution algorithm and the WalkSAT algorithm. PL-Resolution is a sound and complete algorithm that can be used to determine satisfiability and unsatisfiability with certainty. WalkSAT can determine satisfiability if it find a model, but it cannot guarantee to find a model even there exists one. However, WalkSAT is much faster than PL-Resolution, which makes WalkSAT more practical; and we have analysis the performance between these two algorithms, and the performance of WalkSAT is acceptable if the problem is not so hard. In conclusion, if we just want to solve the problem and need it be done very quickly, WalkSAT is the best choice; we also can adjust the argument *max_flips* to high to get more chance of finding the solution. On the other hand, if you have plenty of time, you can chose PL-Resolution which will guarantee give you the answer of whether the sentence is satisfy.

# References

[1] Stuart J. Russell and Peter Norvig. Upper Saddle River, New Jersey 07458, prentice hall, 2010. Artificial Intelligence: A Modern Approach, 3rd ed.

[2] Iwama, K.. CNF-satisfiability test by counting and polynomial average time. SIAM Journal on Computing, 18 (1989), 385-387.

[3] Han, H. Increasing the effectiveness of deduction in propositional SAT solvers. University of Colorado at Boulder). ProQuest Dissertations and Theses, (2011), 123.

[4] Huang, W., & Li, W.. A hopeful CNF-SAT algorithm-- its high efficiency, industrial application and limitation. Journal of Computer Science and Technology, 13 (1998), 9-12.

[5] Johnson, C. R.. The implementation of a DNA computer to solve n-variable 3 CNF SAT problems. University of Southern California). ProQuest Dissertations and Theses, (2004), 129-129.

[6] Qixin Wang, Menghui Li, Li Charlie Xia, Ge Wen, Hualong Zu, Mingyi Gao (2013), Genetic Analysis of Differentiation of T-helper lymphocytes, Genetics and Molecular Research, Vol.12, No.2, PP. 972 - 987

[7] Qixin Wang, Menghui Li, Hualong Zu, Mingyi Gao, Chenghua Cao, Li Charlie Xia(2013), A Quantitative Evaluation of Health Care System in US, China, and Sweden, HealthMED, Vol.7, No.4, PP. 1064-1074

[8] Qixin Wang, Yang Liu, Xiaochuan Pan (2008), Atmosphere pollutants and mortality rate of respiratory diseases in Beijing, Science of the Total Environment, Vol.391 No.1, pp143-148.

[9] Qixin Wang, Chenghua Cao, Menghui Li, Hualong Zu (2013) A New Model Based on Grey Theory and Neural Network Algorithm for Evaluation of AIDS Clinical Trial, Advances in Computational Mathematics and its Applications, Vol.2, No.3, PP. 292-297

[10] Hualong Zu, Qixin Wang, Mingzhi Dong, Liwei Ma, Liang Yin, Yanhui Yang(2012), Compressed Sensing Based Fixed-Point DCT Image Encoding, Advances in Computational Mathematics and its Applications, Vol.2, No.2, PP. 259-262

[11] Hangwei Qian, Chenghua Cao, Li Liu, Hualong Zu, Qixin Wang, Menghui Li, Tao Lin, Exploring the Network Scale-out in virtualized Servers, Proceeding of International Conference on Soft Computing and Software Engineering (SCSE 2013)

[12]Yunfeng Ling, etc., A Fanotype Interference Enhancd Quantum Dot Infrared Photodetector, Vasinajindakaw, Puminun… Applied Physics Letters, 2011.

[13]Yunfeng Ling, Nan Wu, etc., Thin Film Thickness Variation Measurement Using Dual LEDs and Reflectometric Interference Spectroscopy Model in Biosensor, Yunfeng Ling, Nan Wu, etc., SPIE Photonics West, San Francisco, 2010.

[14]Nan Wu, Wenhui Wang, Yunfeng Ling, etc., Label free Detection of Biomolecules Using LED Technology, SPIE Photonics West, San Francisco, 2010.

[15]Yunfeng Ling, etc.,Design of Omnidirectional Vision Reflector based on VC and Matlab, Computer Applications and Software, China, 2008.

[16] Hangwei Qian, Hualong Zu, Chenghua Cao, Qixin Wang (2013), CSS: Facilitate the Cloud Service Selection in IaaS Platforms Proceeding of IEEE International Conference on Collaboration Technologies and Systems (CTS)

[17] Y. Han and A. T. Chronopoulos. Distributed Loop Scheduling Schemes for Cloud Systems. 2013 IEEE 27th International Parallel and Distributed Processing Symposium Workshops & PhD Forum (IPDPSW), Boston, MA, May 2013.

[18] Y. Han and A. T. Chronopoulos. Scalable Loop Self-Scheduling Schemes Implemented on Large-Scale Clusters, 2013 IEEE 27th International Parallel and Distributed Processing Symposium Workshops & PhD Forum (IPDPSW), Boston, MA, May 2013.

[19] Y. Han and A. T. Chronopoulos. A Hierarchical Distributed Loop Self-Scheduling Scheme for Cloud Systems, The 12th IEEE International Symposium on Network Computing and Applications, NCA 2013, Boston, MA, August 2013.

[20] Hangwei Qian, Qixin Wang (2013) Towards Proximity-aware Application Deployment in Geo-distributed Clouds, Advances in Computer Science and its Applications, Vol 2, No 3, PP. PP. 382-386

[21] Liangjun Xie, Nong Gu, Dalong Li, Zhiqiang Cao, Min Tan, and Saeid Nahavandi, "Concurrent Control Chart Patterns Recognition with Singular Spectrum Analysis and Support Vector Machine", Computers & Industry Engineering, Vol 64, I.1, 2013, pp. 280-289.
Image processing

[22] Liangjun Xie, Nong Gu, Ziqiang Cao, Dalong Li, "A Hybrid Approach for Multiple Particle Tracking Microrhelogy", International Journal of Advanced Robotic Systems, Vol. 10, 2013, DOI: 10.5772/54364.

[23]Liangjun Xie, Dalong. Li, Steven J. Simske, "Feature dimensionality reduction for example-based image super-resolution", Journal of Pattern Recognition Research, Vol 2, pp 130-139, 2011.

[24]Hangwei Qian, Elliot Miller, Wei Zhang, Michael Rabinovich, Craig E Wills. Agility in virtualized utility computing. Proceeding Workshop on Virtualization Technologies in Distributed Computing, 2007.

[25]WeiZhang, Hangwei Qian, Craig E Wills, Michael Rabinovich. Agile resource management in a virtualized data center. Proceeding of the First Joint WOSP/SIPEW International Conference on Performance Engineering, 2010.

[26]Hangwei Qian and Michael Rabinovich. Application Placement and Demand Distribution in a Global Elastic Cloud: A Unified Approach. Proceeding of the 10th USENIX International Conference on Autonomic Computing (ICAC), 2013.

[27]Hangwei Qian, Qian Lv. Proximity-aware Cloud Selection and Virtual Machine Allocation in IaaS Cloud Platforms. Proceedings of the IEEE International Workshop on Internet-based Virtual Computing Environment , 2013.